%% file: main.tex
\documentclass{article}

\usepackage{microtype}
\usepackage{graphicx}
\usepackage{subcaption}
\usepackage{booktabs}
\usepackage{xurl}
\usepackage{hyperref}

\usepackage[preprint]{icml2026}

\usepackage{amsmath}
\usepackage{amssymb}
\usepackage{mathtools}
\usepackage{xspace}
\usepackage[capitalize,noabbrev]{cleveref}

\newcommand{\sys}{DeaMoE\xspace}

\icmltitlerunning{DeaMoE: Efficient MoE Structure for Fast Small-Batch Decoding}

\begin{document}

\twocolumn[
  \icmltitle{DeaMoE: Efficient MoE Structure for Fast Small-Batch Decoding}

  \begin{icmlauthorlist}
    \icmlauthor{Zewen Jin}{ustc,iaihf}
    \icmlauthor{Shen Fu}{ustc}
    \icmlauthor{Zeping Duan}{ustc}
    \icmlauthor{Shannon Wang}{iaihf}
    \icmlauthor{Weihao Wu}{ustc}
    \icmlauthor{Chengjie Tang}{iaihf,sxu}\\
    \icmlauthor{Congkun Ai}{ustc}
    \icmlauthor{Ping Gong}{ustc}
    \icmlauthor{Zijian Dai}{ustc,iaihf}
    \icmlauthor{Youhui Bai}{ustc}
    \icmlauthor{Cheng Li}{ustc,iaihf}
  \end{icmlauthorlist}

  \icmlaffiliation{ustc}{University of Science and Technology of China}
  \icmlaffiliation{iaihf}{Institute of Artificial Intelligence, Hefei Comprehensive National Science Center}
  \icmlaffiliation{sxu}{Shanxi University}

  \icmlkeywords{Mixture-of-Experts, Large Language Models, Efficient Inference, Small-Batch Decoding}

  \vskip 0.3in
]

\printAffiliationsAndNotice{}

\begin{abstract}
Mixture-of-Experts (MoE) models have been widely adopted in real-time interactive applications such as coding assistants, real-time audio-video interaction systems. To meet the extremely low response latency requirements of these scenarios, practitioners commonly employ small-batch decoding, under which MoE inference becomes memory-bound and is severely bottlenecked by expert weight loading. However, this bottleneck has received limited attention, and existing solutions such as post-training weight compression or fine-grained expert design during pre-training either degrade model accuracy or introduce additional computation and communication overhead. To tackle this issue, we propose \sys, a decoding-efficient MoE architecture, in which the experts are grouped into several departments, and the experts belonging to the same department share most parameters since they come from the same professional field, and additionally each expert contains a few private parameters to reflect its uniqueness. Moreover, we design customized two-stage routing strategy for \textbf{\sys} to avoid redundant loading, under which \sys greatly improves the efficiency during LLM decoding.
Compared with vanilla MoE, \sys reduces per-step loaded weights by up to 50.9\% and achieves up to 1.33$\times$ end-to-end TPOT speedup for the pre-trained 7B model on A40, and up to 2.00$\times$ and 1.97$\times$ peak speedup for DeepSeek-V3 on A40 and H100 in microbenchmarks.
\end{abstract}

\input{sections/1.introduction}
\input{sections/2.related_work}

\input{sections/3.design}
\input{sections/4.eval.training}

\input{sections/5.eval.inference}

\section{Conclusion}
In this paper, we proposed an efficient DeaMoE structure to improve the performance of MoE in LLM decoding. It was observed that some experts in a MoE layer show a high degree of similarity, indicating parameter redundancy. DeaMoE groups all the experts into several departments. The experts belonging to the same department share most common parameters because they capture shared knowledge within the same functional category. In addition, we proposed a customized two-stage routing strategy. Instead of directly assigning the tokens to experts, the tokens are first allocated to the corresponding departments and then each department uniformly processes all the collected tokens at once and assigns them to the corresponding experts for further differentiated processing. Combining the DeaMoE structure and the routing strategy, \sys delivers consistent decoding speedups in both end-to-end serving and microbenchmarks: for vLLM serving of our pre-trained 7B model, we observe up to 1.33$\times$ end-to-end TPOT speedup on A40, and for DeepSeek-V3, the peak microbenchmark speedups reach 2.00$\times$ on A40 and 1.97$\times$ on H100.

\bibliography{reference}
\bibliographystyle{icml2026}

\input{sections/appendix}

\end{document}

%% file: sections/1.introduction.tex
\section{Introduction}
\label{sec:introduction}
Mixture-of-Experts (MoE) models have become a central approach for scaling large language models (LLMs) by increasing model capacity while keeping per-token computation manageable.
Consequently, MoE architectures have been widely adopted in state-of-the-art LLMs and large-scale industrial deployments~\cite{deepseekV3, Qwen3, kimiK2, longcat}.

It has been witnessing that MoE models are increasingly deployed in real-time and interactive applications such as code completion, voice assistants, and multi-modal audio-video systems, where strict latency requirements make decoding efficiency a primary concern. In these scenarios, practitioners typically favor small-batch decoding to balance the trade-off between decoding latency and throughput, as larger batch sizes improve throughput but introduce unacceptable delays. However, under small-batch decoding, MoE inference becomes memory-bound and is severely bottlenecked by expert weight loading. During autoregressive decoding, only a few tokens are processed per step while multiple experts are activated per token, leading to scattered expert access with minimal weight reuse. As a result, decoding latency is dominated by repeatedly loading expert parameters rather than computation.

Despite its importance in real-time serving, this small-batch expert loading bottleneck has been largely overlooked by existing MoE optimizations. While post-hoc expert compression~\cite{molae,mobe,moe-svd,d2moe} and pre-training-time architectural modifications with more fine-grained experts~\cite{nemotron} may indirectly reduce expert footprints, they either risk degrading model quality or rely on serving-time parallelism that is often unavailable in small-batch settings. Consequently, no existing approach directly targets the expert loading inefficiency that dominates small-batch MoE decoding, motivating the need for decoding-driven designs that explicitly reduce per-step expert loading without sacrificing model quality.

\begin{figure}[!t]
  \centering
  \includegraphics[width=0.94\columnwidth]{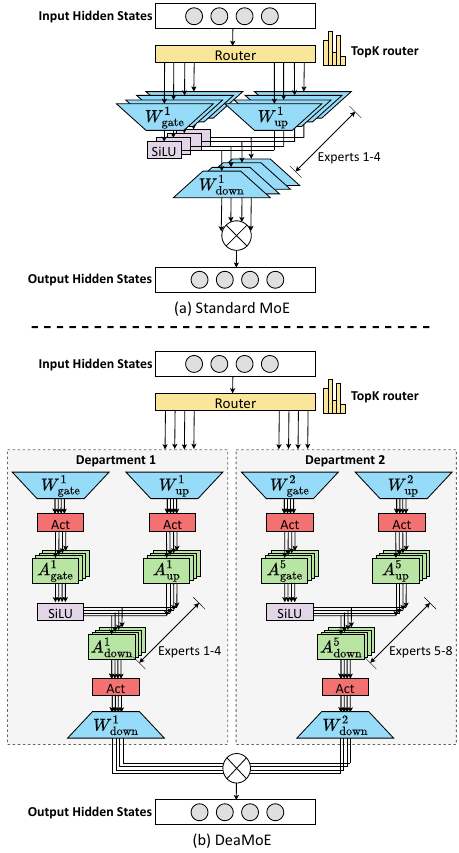}
  \caption{Architecture of \sys and standard MoE.}
  \label{fig.moe.structure}
\end{figure}

To this end, we propose \sys, a new MoE architecture that reduces expert weight loading overhead while preserving model quality under comparable parameter and FLOPs budgets.
As \cref{fig.moe.structure}(b) shows, our key idea is to group experts into a small number of \emph{departments}: experts within the same department share most parameters as a common backbone, while each expert retains a small set of private parameters to capture its uniqueness.
To avoid redundant loading during decoding, we further design a customized two-stage routing strategy that first selects departments and then routes tokens to experts within the selected departments, improving weight reuse across tokens within each decoding step.

We summarize our main contributions as follows:

\textbf{(1)} We propose \sys, a decoding-efficient MoE architecture which groups experts into several departments. The experts belonging to the same department share most parameters and each expert only has a small number of private parameters to distinguish  it from others. Such a structure greatly reduces the experts' parameter redundancy.

\textbf{(2)} We  design a customized two stage routing strategy for DeaMoE. We first route all the tokens to their corresponding departments and each department will uniformly process all the collected tokens at once. Then the department will assign the tokens to the corresponding experts for further differentiated processing. Using such a routing strategy, \sys can truly demonstrate its performance advantages during the decoding stage.

\textbf{(3)} We pre-train a 7.3B  \sys model and its budget-matching standard MoE counterpart over 110B tokens. \sys preserves model quality on pre-training loss and a dozen downstream benchmarks.

\textbf{(4)} We conduct extensive experiments demonstrating that \sys delivers clear advantages in small-batch decoding scenarios under both Ampere and Hopper GPUs. For end-to-end vLLM serving of the 7.3B model on A40, \sys achieves up to 1.33$\times$ speedup in TPOT, and for DeepSeek-V3, it reaches peak microbenchmark speedups of 2.00$\times$ on A40 GPU and 1.97$\times$ on H100 GPU.

%% file: sections/2.related_work.tex
\section{Background and Motivation}
\label{sec:background}

\subsection{Standard MoE Structure}
As shown in \cref{fig.moe.structure}(a), the MoE layer typically consists of a router and multiple individual experts. Each expert $i$ includes three projection functions, which are gate projection $g_i(x) = \mathrm{SiLU}(W^{i}_{\mathrm{gate}}\mathbf{x})$, up projection $u_i(x) = W^{i}_{\mathrm{up}}\mathbf{x}$, and down projection $d_i(x) = (W^{i}_{\mathrm{down}})^Tx $, where $W^{i}_{\mathrm{gate}}, W^{i}_{\mathrm{up}}, W^{i}_{\mathrm{down}} \in \mathbb{R}^{h_{\mathrm{ffn}}\times h}$. Given a token $x$, expert $i$ computes its output by
\begin{equation}
\label{eq:standard-moe-ffn}
\mathrm{MoE}_{i}(\mathbf{x})
=d_i(g_i(x) \odot u_i(x))
\end{equation}
The MoE router first assigns each token to its best determined top-k experts out of $n$ experts by
\begin{equation}
\label{eq:routing}
\mathcal{S}(\mathbf{x}) = \operatorname{TopK}\left(W_{r}\mathbf{x}, k_\mathrm{expert}\right),
\end{equation}
and then computes the output as the linearly weighted combination of each expert’s computation on the token
\begin{equation}
\label{eq:moe-output}
\mathrm{MoE}(\mathbf{x})
=
\sum_{i\in \mathcal{S}(\mathbf{x})}
p_i(\mathbf{x})\, \mathrm{MoE}_i(\mathbf{x}),
\end{equation}
where the weighted vector $p_i$ are routing weights after softmax normalization and scaling.
\subsection{Expert Loading Issue in Small-batch Decoding}

MoE models have been widely deployed in real-time and interactive applications, where strict responsiveness requirements impose highly demanding constraints on decoding latency. Representative scenarios include real-time code completion and interactive programming assistants, voice assistants with streaming ASR(Automatic Speech Recognition) and TTS(Text-to-Speech), and real-time audio-video interaction systems such as multimodal agents for video conferencing, live translation, and embodied AI. In these settings, users typically expect token-level response latencies on the order of tens of milliseconds, making decoding latency a primary optimization target.

However, MoE inference exhibits a well-known trade-off among batch size, decoding latency, and decoding throughput~\cite{moe-sc}. On the one hand, large batch sizes significantly improve throughput by amortizing computation and memory access costs, but they also introduce prohibitive latency that violates real-time constraints. On the other hand, reducing the batch size leads to much lower latency but severely degrades throughput due to underutilized compute resources. Consequently, for latency-critical scenarios, both prior work and industrial deployments commonly adopt small-batch decoding regimes, typically in the range of 4–128 tokens per step, which strike a practical balance between ultra-low latency and acceptable throughput~\cite{moe-cap}.

Unfortunately, small-batch MoE inference poses a severe expert loading bottleneck during autoregressive decoding, where the majority of decoding time is dominated by loading expert parameters rather than computation. This is because during each decoding step, only a small number of tokens are processed, while the router often activates multiple experts per token. The activated experts are scattered across many distinct expert indices, with very limited reuse of expert weights across tokens within the same step. In this regime, MoE decoding is no longer compute-bound by matrix multiplications, but instead becomes memory-bound, as expert parameters must be repeatedly fetched from memory, leading to high effective memory traffic and increased latency~\cite{deepseekV3, deepseek-insight, moe-sc}.

This issue is particularly evident in large-scale MoE models. For example, each expert in DeepSeek-V3 consists of three matrices of shape $[7168, 2048]$, amounting to approximately 44 MB of FP8 weights, which is already close to the upper limit of the H100 GPU’s L2 cache capacity~\cite{deepseekV3}. As modern MoE models continue to scale in both the number and size of experts to improve capacity and specialization, the small-batch expert loading problem becomes even more pronounced. Recent models such as Kimi-K2, Longcat-Omni, and Qwen3-Next adopt 384, 512, and 512 routed experts, respectively~\cite{kimiK2,longcat,Qwen3Next}. With hundreds of experts available, the likelihood that different tokens within a small batch select the same expert further diminishes, exacerbating parameter loading redundancy and increasing memory bandwidth pressure per decoding step.

\subsection{Existing Methods and Their Limitations}
\label{sec:background:related_work}

In this paper, we focus on reducing \emph{small-batch decoding} latency for MoE models, with an emphasis on mitigating the \emph{expert weight-loading} bottleneck.
Despite its practical importance in interactive serving, most existing MoE optimizations have largely been developed for objectives other than small-batch decoding (e.g., model compactness, training scalability, or large-batch throughput), leaving the decoding-time expert loading issue relatively underexplored.

Given this gap, we review prior approaches that are not explicitly designed for small-batch decoding but could, in principle, be related to our target.

\paragraph{Post-hoc expert compression.}
This line of work compresses a \emph{trained} MoE model by reducing the footprint of expert weights.
SVD-style approaches (e.g., MoLAE, D$^2$MoE, and MoE-SVD) factorize expert weight matrices into low-rank components, typically by introducing shared bases and expert-specific coefficients, so that each expert can be represented using a small number of factors~\cite{molae,d2moe,moe-svd}. MoBE further studies cross-expert redundancy by constructing a shared expert basis and learning expert-specific mixing on top of it, aiming to improve compression-quality trade-offs compared with purely low-rank factorization~\cite{mobe}. While such methods can substantially reduce total parameters, they may introduce approximation constraints that affect model quality, and the decoding-time benefit can be limited when small batches still activate many distinct experts, resulting in considerable per-step weight movement.

\paragraph{Pre-training-time structural changes.}
Another direction modifies the MoE architecture \emph{during pre-training} to reduce the per-expert footprint, often with the expectation that, under sufficiently large expert parallelism, each GPU handles fewer expert parameters~\cite{nemotron}. These approaches typically preserve model quality better than post-hoc compression, but their latency benefit can diminish when the serving-time cluster size and request volume constrain achievable parallelism, particularly in the small-batch decoding regime.
Overall, existing compression methods and pre-training-time structural changes provide useful insights, but neither explicitly targets the key bottleneck we study—\emph{expert weight loading under small-batch decoding}. This motivates our decoding-driven design that directly reduces per-step expert loading while preserving model quality under comparable parameter and FLOPs budgets.

\subsection{Motivation}
\label{sec:background:motivation}
Unlike the above expert compression work, we aim to propose a \emph{more loading-efficient parameterization}. Rather than making experts smaller, we make expert weights \emph{more reusable} across tokens and across experts during decoding.
Our key observation is that modern MoE scaling increasingly relies on a large expert pool, yet trained experts often exhibit substantial redundancy. Empirically, multiple experts can learn similar transformations (e.g., high cross-expert similarity in representations or weights), indicating that a considerable portion of expert parameters corresponds to shared ``backbone'' knowledge, while true specialization may reside in a smaller set of expert-specific components~\cite{Discovery}. This suggests an opportunity to factor out and reuse the shared parts, reducing redundant loading without forcing aggressive per-expert compression.

At the same time, prior experience indicates that post-hoc modifications on a trained MoE (e.g., compression or re-parameterization) can be brittle and may degrade model quality. To avoid such issues, we take a \emph{pre-training-first} approach: we redesign the MoE structure to explicitly encode reuse, and pre-train the model from scratch under the new parameterization. This results in \sys, which organizes experts into departments with shared large backbones and lightweight expert-specific transforms.
With \sys, small-batch inference can reduce per-step weight movement and thus further lower decoding latency. This brings two practical benefits: (i) better meeting \emph{ultra-low latency} requirements in interactive applications, and (ii) under a fixed latency budget, enabling a larger feasible batch size and hence higher throughput.

%% file: sections/3.design.tex
\section{DeaMoE: New Model Structure}
\label{sec:method}
\label{sec:method:architecture}

In this section, we propose \sys to address the above mentioned decoding inefficiency of MoE.
\subsection{Main Structure}
Unlike the standard MoE in which all the experts are totally individual (\Cref{fig.moe.structure}(a)), in \sys, the experts are grouped into several departments (\Cref{fig.moe.structure}(b)).
The experts from different departments are individual since their areas of responsibility are completely different, while the experts from the same department share most parameters since they are responsible for different sub-disciplines within the same professional field.
Specifically, in each department, all the experts share the large gate, up, and down projection matrices, and additionally each expert $i$ introduces three sub-matrices $A^i_g$, $A^i_u$, and $A^i_d$ (all $\in \mathbb{R}^{h_{\mathrm{ffn}}\times h_{\mathrm{ffn}}}$) to distinguish it from other experts.
Under this scheme, the gate projection function of an expert $i$ belonging to department $j$ is expressed as
\begin{equation*}
    \hat{g}_i(x) =  \mathrm{SiLU}(A^i_g\sigma({W}^{j}_gx)),
\end{equation*}
where ${W}^{j}_g$ is a common gate projection weight of department $j$, and $A^i_g$ is private gate projection weight of expert $i$. Similarly, the up and down projection functions become
\begin{equation*}
    \hat{u}_i(x) =  A^i_u\sigma({W}^{j}_ux),
\end{equation*}
and
\begin{equation*}
    \hat{d}_i(x) =  ({W}^{j}_d)^T\sigma((A^i_d)^Tx).
\end{equation*}
Here $\sigma(\cdot)$ is an activation function which is necessary to improve the stability of the training process. We investigate the effect of different activation functions in \Cref{sec:ablation}, including removing this activation function.
Then a \sys expert $i$ from department $j$  computes a given token $x$ by
\begin{equation}
\label{eq:dea-moe-ffn}
\mathrm{DeaMoE}_{i}(\mathbf{x})
=\hat{d}_i(\hat{g}_i(x) \odot \hat{u}_i(x))
\end{equation}

\subsection{DeaMoE Routing}
In DeaMoE, all $n$ experts are first grouped into $m$ departments. We still use the router function (\ref{eq:routing}) to determine the assigned experts for a given token and use function (\ref{eq:moe-output}) to obtain the final output of the DeaMoE layer. Rather than directly sending the token to its corresponding expert computing its output by (\ref{eq:dea-moe-ffn}), we first collect all the tokens belonging to the experts from the same department $j$,  denoted by $X^j$, and compute $X^j_g = \sigma(W_g^jX^j)$, $X^j_u = \sigma(W^j_uX^j)$. Then $X^j_g$ and $X^j_u$ will be allocated to the corresponding experts for further computing with their private parameters $A^i_g$, $A^i_u$, and $A^i_d$. Let $X_g^{j,i}$ and $X_u^{j,i}$ denote the data from $X^j_g$ and  $X^j_u$ belonging to expert $i$.
Then expert $i$ computes
\begin{equation*}
    Y^{j,i} = \mathrm{SiLU}(A^i_g X_g^{j,i})  \odot  (A^i_u X_u^{j,i}),
\end{equation*}
and
\begin{equation*}
     Z^{j,i} = \sigma((A^i_d)^TY^{j,i}).
\end{equation*}
 Similarly, we collect all the $Z^{j,i}$ for the same department $j$, denote as $Z^j$, and compute the expert output $O^j = (W_d^j)^T Z^j$. Compared with directly using function (\ref{eq:dea-moe-ffn}), such a manner greatly reduces repeated loading of the department parameters $W^j_g, W^j_u,$ and $W^j_d$, which is of great importance in LLM decoding.

\begin{table*}[t]
\caption{Baseline MoE models and their budget-matched \sys configurations.}
\label{table:models.and.deamoe}
\centering
\small
\setlength{\tabcolsep}{2pt}
\begin{tabular}{lcccc|cccc|cc}
\toprule
 & \multicolumn{4}{c}{Baseline Config} & \multicolumn{4}{c}{DeaMoE Config} & \multicolumn{2}{c}{Comparison}  \\
\cmidrule(lr){2-5} \cmidrule(lr){6-9} \cmidrule(lr){10-11}
MoE Models &$k_\mathrm{expert}$& $h$ & $h_{\mathrm{ffn}}$ &
$n_\mathrm{expert}$& $n_{\mathrm{dept}}$ & $n_\mathrm{expert}$ & $k_{\mathrm{dept}}$ & $k_{\mathrm{expert}}$
& FLOPs ($\times$) & Params ($\times$) \\
\midrule
DeepSeek-V3     &8 & 7168 & 2048 & 256 & 8  & 864  & 5 & 10 & 0.995 & 0.996 \\
DeepSeek-V2    & 6  & 5120 & 1536 & 160& 8  & 512  & 5 & 10 & 1.010 & 1.010 \\
Qwen3-235B-A22B      &8    & 4096 & 1536 & 128& 8  & 320  & 5 & 8  & 1.005 & 1.000 \\
Kimi-K2      &8    & 7168 & 2048 & 384& 16 & 1280 & 5 & 10 & 1.000 & 0.994 \\
Baseline-7B &8  & 1024 & 384  & 256& 8  & 672  & 4 & 10 & 1.014 & 1.017 \\
\bottomrule
\end{tabular}
\end{table*}

\subsection{Comparing \sys vs Standard MoE Counterpart}
\label{sec:method:ratio}
In this section, we compare \sys against the standard MoE counterpart in terms of the expert loading amount.
As mentioned before, for a fair comparison, we consider the same budgets of both total parameter count and per-token FLOPs.
We summarize the selected MoE baselines and their budget-matched \sys counterparts in \Cref{table:models.and.deamoe}~\cite{deepseekV3,Qwen3,kimiK2}.
Baseline-7B is the model we pre-train and use for a detailed accuracy performance comparison with \sys in \Cref{sec:evluation:pretrain}.
For each baseline, we search for a \sys configuration that respects the prescribed budget constraints and maximizes the reduction in expert weight loading under small-batch decoding.
Concretely, we use group top-$k$ routing to cap the activated FLOPs per token associated with department matrices~\cite{deepseekV3}.
The budget accounting includes both the FFN parameters and computation as well as the MoE router itself.

\begin{figure}
  \centering
    \centering
    \includegraphics[width=\columnwidth]{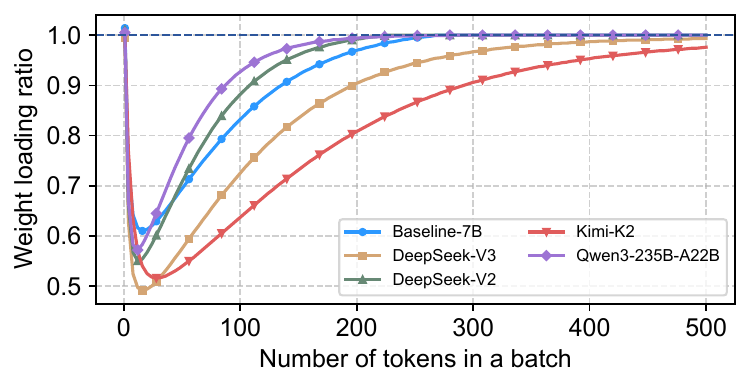}
    \caption{
    Comparison of weight loading for various MoE models after their adaptation to the DeaMoE architecture, normalized to their original MoE baselines (horizontal line at 1.0) w.r.t the batch size.}
    \label{fig.loading.ratio}
\end{figure}

\Cref{fig.loading.ratio} illustrates how the expected amount of expert parameters loaded per decoding step changes with the number of tokens, normalized by the baseline model under the same configuration.
The loading ratio exhibits a non-monotonic trend: it decreases at small token counts and then increases as the batch grows.
When the batch is very small, cross-token weight sharing is limited and the normalized loading remains relatively high.
As the number of tokens increases from this regime, reuse improves and the normalized loading decreases. With sufficiently large batches, the union of activated departments/experts expands and covers more distinct experts, so the loaded weights per step approach those of the baseline and the ratio increases toward 1.0.
Consistent with this behavior, \sys achieves the largest reduction in weight loading in the practical small-batch regime.
For the configuration matched to DeepSeek-V3, the loading ratio stays between 49.1$\%$ and 62.8$\%$ when the number of tokens is between 4 and 64, indicating substantially less expert weight movement per decoding step than the baseline.

\subsection{Recipes for \sys Training}
\label{sec:method:activation}
In this section, we analyze how some decision choices affect the model quality in pre-training.
The related ablation study is shown in \Cref{sec:ablation}.

\paragraph{Non-linear operator between department and expert projections.}
We insert non-linear operators between the department-level backbones and the expert-specific transforms to avoid stacking consecutive linear maps, which can hurt model quality.
We consider several choices, including SiLU, GeLU, and RMSNorm~\cite{silu, gelu, rmsnorm}.
In \sys, we finally adopt SiLU as a smooth activation and apply it consistently at all three interfaces (gate/up/down) to modulate both shared and expert-specific features while preserving signal diversity.
Compared to SiLU, replacing it with GeLU or RMSNorm, or removing the non-linear operator altogether, results in higher training loss in \Cref{sec:ablation}.

\paragraph{Initialization of the expert parameters.}
In \sys, each expert is implemented as a lightweight transform on top of a shared department backbone. Under this factorized design, the initialization of expert matrices is critical.
Following common practice in industry~\cite{switch-transformer}, standard MoE models initialize \emph{expert} weights with a zero-mean Gaussian scheme, e.g., \texttt{torch.nn.init.normal\_(mean=0, std=$\sigma$)}.
However, in \sys, using such initialization for both \emph{expert} and \emph{department} matrices introduces arbitrary deviations in the backbone representations at the start of training, which can distort the routed features and lead to degraded model quality.

To avoid this issue, we initialize the expert-specific matrices with identity mappings using \texttt{torch.nn.init.eye\_}.
This makes each expert behave equivalently to its corresponding department backbone at initialization, so the model starts from a well-conditioned solution with no artificial specialization.
Expert-specific behavior is then gradually learned through training rather than being imposed by random noise.
Our ablation results in \Cref{sec:ablation} show that identity initialization achieves lower training loss compared to random initialization, validating the importance of a near-identity starting point for expert transforms.

\begin{table*}[!t]
\centering
\small
\setlength{\tabcolsep}{4pt}
\caption{Accuracy results with downstream tasks for MoE models pre-trained with 110B tokens.}
\label{table.eval.downstream.acc}
\begin{tabular}{c|cccccccccc}
\toprule[0.8pt] \textbf{Model}
         & \textbf{BoolQ}& \textbf{PIQA}  & \textbf{SIQA}   & \textbf{HSwag} & \textbf{WinoG} & \textbf{RaceH} & \textbf{AnliR1} & \textbf{AnliR2} & \textbf{AnliR3} & \textbf{OBQA} \\
\midrule
\sys-7B & \textbf{62.39}& 73.39          & \textbf{42.07}  & 57.04 & \textbf{58.01} & 39.62 & \textbf{32.80}  & \textbf{34.3}   & \textbf{35.25}  & 34.8 \\
Baseline-7B & 61.47& \textbf{74.43} & 41.97           & \textbf{57.54} & 55.25 & \textbf{39.71} & 32.0   & 31.9            & 34.17           & \textbf{37.6} \\
\bottomrule
\end{tabular}
\end{table*}

\paragraph{Attempt to enforce $k_\mathrm{dept}$ coverage of departments per token.}
Under group-limited top-$k$ routing, $k_\mathrm{dept}$ serves as an upper bound on the number of departments activated by each token, so that the department-side FLOPs remain capped for a budget-matched comparison with the baseline. It is not a requirement that every token must cover exactly $k_\mathrm{dept}$ departments. If the router selects experts from fewer departments, the realized computation is simply below this matched-FLOPs upper bound, which is acceptable as long as model quality is preserved.

For completeness, we evaluate a strict variant that enforces exact department coverage. After the router selects the top-$k$ experts, we count the number of distinct departments they span. If fewer than $k_\mathrm{dept}$ departments are covered, we replace the lowest-ranked selected expert with the highest-probability expert from an uncovered department, until the final selection spans exactly $k_\mathrm{dept}$ departments.

However, this strict enforcement degrades model quality in practice. Our ablation results (\Cref{sec:ablation}) show that forcing tokens to activate experts from additional departments perturbs the learned routing preferences and harms specialization. These results suggest that exact department coverage is not a desirable objective; the original soft group-limited top-$k$ constraint better preserves the quality--efficiency trade-off.

%% file: sections/4.eval.training.tex
\section{Accuracy Evaluation after Pre-Training}
\label{sec:evluation:pretrain}

In this section, we compare \sys against a standard MoE baseline by reporting their performance on downstream benchmarks and language modeling perplexity after pre-training.

\subsection{Pre-Training}
For a fair comparison, we pre-train both MoE models with an identical size of 7.3B parameters.
The baseline adopts a conventional MoE feed-forward structure following the hyper-parameter configuration listed in \Cref{table:models.and.deamoe}, while \sys is configured to match the baseline in terms of total parameter count and per-token FLOPs.
\Cref{table:models.and.deamoe} summarizes the MoE-related configurations of both models.
We set the hidden size $h$ and feed-forward dimension $h_\mathrm{ffn}$ to preserve a ratio of $\frac{h}{h_\mathrm{ffn}}=2.67$, consistent with Qwen3-235B-A22B.
We leverage PyTorch FSDP to train the baseline and \sys~\cite{fsdp}.
The code for training will be open-sourced later, which contains both forward and backward passes.

For both experiments, we use the RedPajama-v1 dataset for pre-training~\cite{RedPajamaV1}.
The models are pre-trained with the same corpus of 110B tokens
with identical software and hardware configurations.
In addition, we keep the optimizer settings and training hyper-parameters the same across both models.

\subsection{Downstream Task Performance}

We then evaluate the downstream task performance of \sys and the baseline MoE model after pre-training. Following common practice, we report accuracy results on a suite of reasoning and question answering benchmarks, including commonsense reasoning, natural language inference, and multi-choice QA tasks, as well as perplexity results on standard language modeling datasets.

\begin{table}[!t]
\centering
\caption{Perplexity results for different MoE models pre-trained with 110B tokens. The lower number is better.}
\label{table.eval.downstream.ppl}
\begin{tabular}{c|ccc}
\toprule[0.8pt] \textbf{Model}
         & \textbf{PTB}  & \textbf{WikiText-103}  & \textbf{WikiText-2}\\
\midrule
\sys-7B & \textbf{61.91}          & 20.31 & \textbf{20.33} \\
Baseline-7B & 64.33          & \textbf{20.18} & 20.43 \\
\bottomrule
\end{tabular}
\end{table}
\Cref{table.eval.downstream.acc} summarizes the accuracy results on ten downstream classification and reasoning tasks.
Overall, \sys achieves performance comparable to the Baseline-7B across all evaluated benchmarks, and outperforms the baseline on six tasks, including SIQA, BoolQ, WinoGrande, and all three ANLI subsets (R1, R2, and R3), indicating comparable model quality compared with the standard MoE architecture on commonsense and natural language inference tasks~\cite{SIQA,boolq,WinoGrande,ANLI}.
Although the baseline slightly outperforms \sys on PIQA, HellaSwag, RaceH and OpenbookQA, the performance gaps remain small, suggesting no degradation in overall task generalization~\cite{PIQA,HellaSwag,race,OpenBookQA2018}.

\Cref{table.eval.downstream.ppl} reports perplexity results on PTB, WikiText-103, and WikiText-2~\cite{PTB,WikiText}.
\sys achieves lower perplexity on PTB and WikiText-2, and remains comparable to the baseline on WikiText-103, demonstrating that the proposed architecture preserves language modeling quality under the same pre-training budget and inference FLOPs consumption.

%% file: sections/5.eval.inference.tex
\section{Evaluation for Inference Speedups}
\label{sec:evluation:inference}

    In this section, we evaluate the decoding efficiency of \sys compared to the standard MoE baseline. We report the performance gains of \sys in end-to-end serving scenarios and per-layer microbenchmarks.

\paragraph{Implementation.}

For model serving, we integrate \sys into vLLM v0.13.0~\cite{vllm-paged-attn}, utilizing Triton~\cite{triton} to implement our customized operators.
To ensure a strictly fair comparison, our kernel implementation aligns closely with vLLM’s highly optimized native \texttt{fused\_moe\_kernel}.
Specifically, we reuse the exact Grouped GEMM primitives employed by the baseline, modifying only the token indexing logic to accommodate the hierarchical department-expert structure.
Beyond kernel-level parity, we also enable vLLM’s CUDA Graph execution for \sys and all baselines. Since decoding is inherently latency-sensitive and typically operates in the small-batch regime, adopting CUDA Graph is widely used in industry to reduce kernel launch overhead.
By rigorously matching both the arithmetic kernels and the runtime configurations, we ensure that the observed speedups are attributed solely to the architectural efficiency of \sys.

\subsection{End-to-End Speedup of Pre-Trained Model}

\begin{figure}[!t]
    \centering
    \includegraphics[width=\columnwidth]{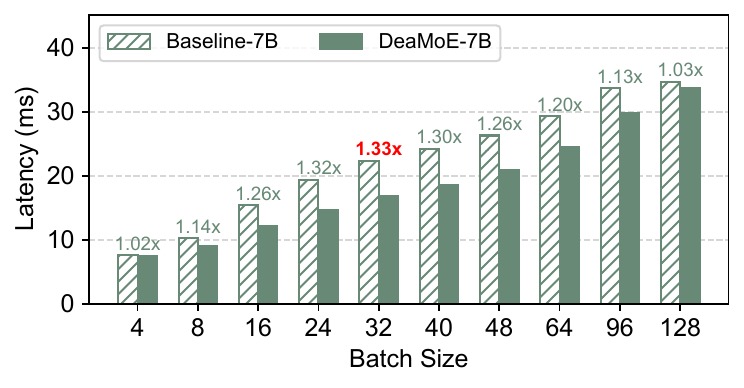}
    \caption{Comparison of average TPOT latency between Baseline-7B and \sys across varying request concurrency on an A40 GPU (\sys's speedups are annotated).}
    \label{fig.end2end.speedup}
\end{figure}

We first evaluate the end-to-end inference performance with Baseline-7B and the \sys counterpart on a single NVIDIA A40 GPU based on vLLM.
To simulate realistic serving dynamics, we generate a continuous stream of requests and measure the average \emph{Time Per Output Token} (TPOT) across varying concurrency levels, which are reported in \Cref{fig.end2end.speedup}.

\begin{figure*}[!t]
    \centering
    \begin{subfigure}{0.33\textwidth}
        \centering
        \includegraphics[width=\linewidth]{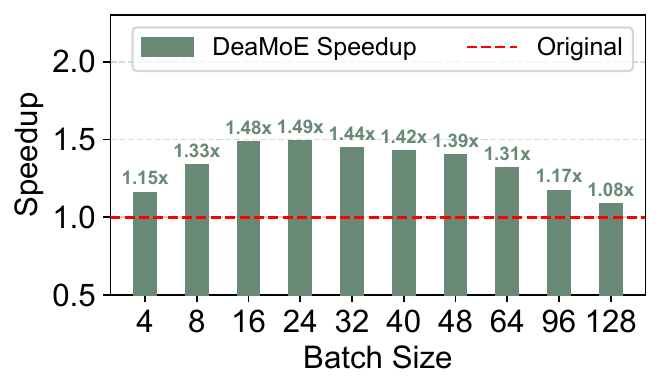}
        \caption{Baseline-7B}
    \end{subfigure}
    \hfill
    \begin{subfigure}{0.33\textwidth}
        \centering
        \includegraphics[width=\linewidth]{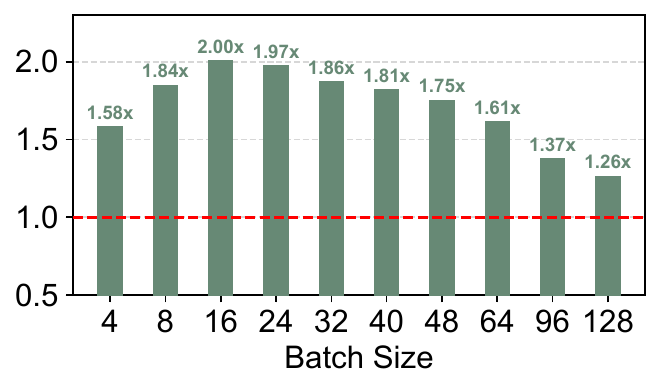}
        \caption{DeepSeek-V3}
    \end{subfigure}
    \hfill
    \begin{subfigure}{0.33\textwidth}
        \centering
        \includegraphics[width=\linewidth]{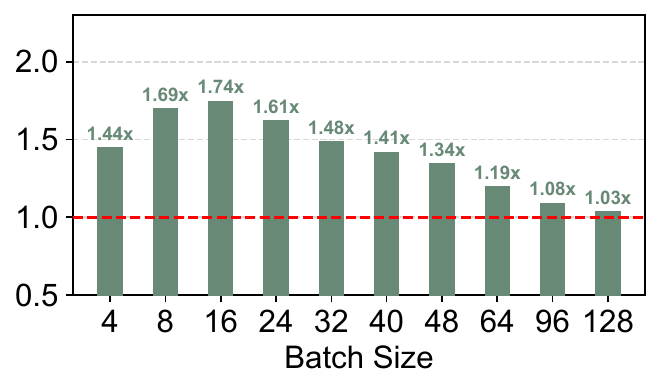}
        \caption{Qwen3-235B-A22B}
    \end{subfigure}

    \caption{Speedup comparison for different models on A40.}
    \label{fig.microbench.speedup.a40}
\end{figure*}

\begin{figure*}[!t]
    \centering
    \begin{subfigure}{0.33\textwidth}
        \centering
        \includegraphics[width=\linewidth]{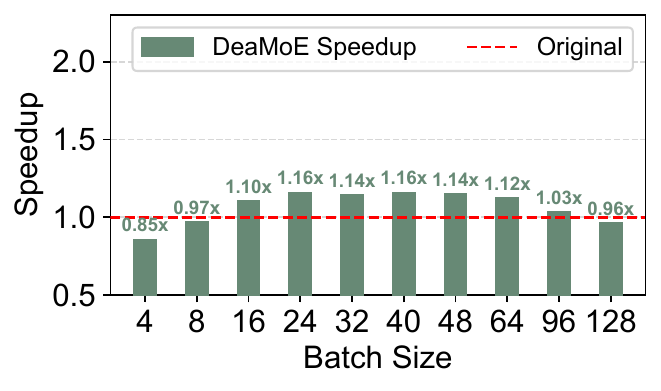}
        \caption{Baseline-7B}
    \end{subfigure}
    \hfill
    \begin{subfigure}{0.33\textwidth}
        \centering
        \includegraphics[width=\linewidth]{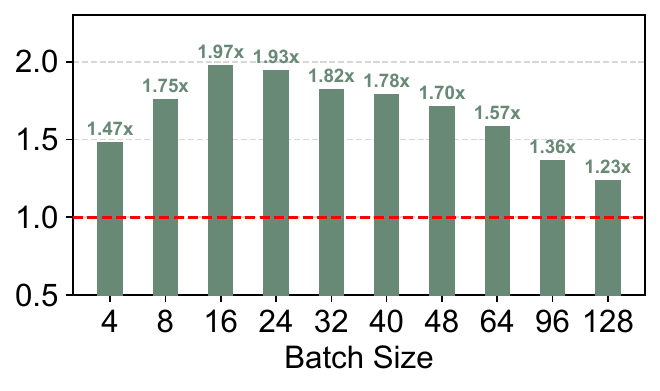}
        \caption{DeepSeek-V3}
    \end{subfigure}
    \hfill
    \begin{subfigure}{0.33\textwidth}
        \centering
        \includegraphics[width=\linewidth]{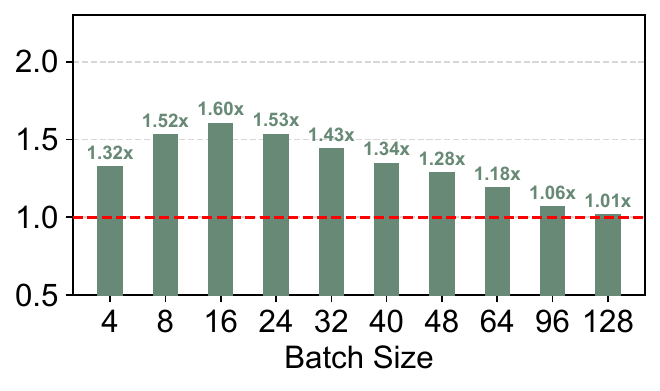}
        \caption{Qwen3-235B-A22B}
    \end{subfigure}

    \caption{Speedup comparison for different models on H100.}
    \label{fig.microbench.speedup.h100}
\end{figure*}

\paragraph{Latency comparison under same batch sizes.}
As shown in \Cref{fig.end2end.speedup}, \sys delivers a latency speedup of 1.02$\times$--1.33$\times$ across the evaluated batch sizes.
Consistent with \Cref{fig.loading.ratio}, the observed weight-loading reduction exhibits a non-monotonic pattern.
Accordingly, the speedup increases as the batch size grows from 4 to 32 (peaking at 1.33$\times$), and then gradually decreases, approaching 1.03$\times$ at the batch size of 128.

\paragraph{Throughput comparison under same latency budget.}
In practical serving, the maximum batch size is often limited to satisfy strict service-level objectives (SLO), particularly on TPOT.
Such latency constraints directly limit achievable throughput.
As shown in \Cref{fig.end2end.speedup}, under a TPOT budget of 30 ms, the baseline and \sys sustain batch sizes of 64 and 96, respectively, corresponding to a 1.50$\times$ throughput improvement.
Under a stricter 20 ms TPOT budget, the supported batch sizes become 24 and 44, respectively, translating to a 1.83$\times$ throughput improvement.

This performance gain confirms that our architecture effectively reduces the memory traffic per decoding step, thereby alleviating the bandwidth bottleneck on the A40 GPU.
\sys sustains this latency and throughput advantage with small-batch scenarios, validating that the reduction in loaded weights directly translates into measurable decoding speedups.

\subsection{Microbenchmarks}

To demonstrate the scalability and hardware universality of \sys, we conduct microbenchmarks on three models ranging from the 7.3B pre-trained baseline up to the DeepSeek-V3 (671B).
To comprehensively assess performance across different hardware generations, we benchmark on both NVIDIA A40 and H100 GPUs.
We measure the per-layer execution time specifically for decoding steps, aiming to capture the pure computation and memory access latency of the MoE layer.
We vary the batch size from 4 to 128 to cover the typical workload range of interactive decoding.

\cref{fig.microbench.speedup.a40} and \cref{fig.microbench.speedup.h100} report the microbenchmark speedups of \sys over the standard MoE baselines on A40 and H100 GPUs across a wide range of batch sizes. Several consistent trends can be observed.

First, on the A40 GPU (\cref{fig.microbench.speedup.a40}), \sys achieves clear and stable speedups for all three models.
The improvement is most pronounced for large-expert models such as DeepSeek-V3 and Qwen3-235B-A22B, with peak speedups approaching $2.00\times$ and $1.74\times$, respectively.
For the smaller Baseline-7B model, \sys still provides noticeable gains.
These results indicate that, on bandwidth-limited hardware, reducing expert weight movement directly translates into lower decoding latency, and the benefit scales with the size of activated experts.

On the H100 GPU (\cref{fig.microbench.speedup.h100}), \sys continues to deliver substantial speedups for DeepSeek-V3 and Qwen3-235B-A22B, demonstrating that the proposed design remains effective on more advanced hardware.
However, the gains for Baseline-7B are noticeably smaller and even show mild regressions at very small and very large batch sizes.
This contrast can be attributed to the interaction between model scale and the memory hierarchy of H100: for small-expert models, expert weights are more likely to fit in and be reused from the large L2 cache, which reduces the baseline’s weight-loading cost and limits the headroom for further optimization.

Overall, these results highlight that the effectiveness of \sys is strongly correlated with the dominance of expert weight loading in the decoding pipeline.
When expert matrices are large and sparsely reused, as in modern large-expert MoE models, \sys yields consistent and significant speedups across both GPU generations. In contrast, for small-expert configurations on high-bandwidth hardware, the bottleneck shifts away from weight movement, and the relative benefit becomes less pronounced. This reinforces our design motivation: \sys is particularly well suited for accelerating small-batch decoding of large-scale MoE models, where expert loading constitutes a primary performance bottleneck.

%% file: sections/appendix.tex
\clearpage
\appendix
 \onecolumn
\section{Ablation Study}
\label{sec:ablation}
To validate the design choices discussed in \Cref{sec:method:activation}, we conduct a controlled ablation study.
Considering the limited resource budget for training, we train a smaller 5B-scale model with 12.4B tokens of RedPajama-V1.
We implement multiple variants that differ only in the corresponding design choice, while keeping the remaining architecture and training recipe unchanged.
The results are summarized in \Cref{table.ablation}, where \emph{Loss} reports the average training loss over the last 200 optimization steps (corresponding to 210M tokens).
Variant No.1 is used in \Cref{sec:evluation:pretrain}.

\begin{table}[h]
\centering
\small
\caption{Training loss with different design choices regarding non-linear operators, expert weight initialization, and department Top-$k$.
}
\label{table.ablation}
\begin{tabular}{l|ccc|c}
\toprule
\textbf{Variant} & \textbf{Operator} & \textbf{Init.} & \textbf{Dep. Top-$k$} & \textbf{Loss} \\
\midrule
No.1*           & SiLU                         & Identity      & Soft                     & 2.504         \\
No.2           & RMSNorm                      & Identity      & Soft                     & 2.510         \\
No.3           & GeLU                         & Identity      & Soft                     & 2.536         \\
No.4           & None                         & Identity      & Soft                     & 2.508         \\
\midrule
No.5           & SiLU                         & Random        & Soft                     & 2.530         \\
\midrule
No.6           & SiLU                         & Identity      & Strict                   & 2.519        \\
\bottomrule
\end{tabular}
\end{table}
\paragraph{Non-linear operator between department and expert projections.}
As shown with variants from No.1 to No.4, under the same training setup, the last-200-step average loss ranks SiLU as the best-performing operator among the evaluated variants, with GeLU, RMSNorm, and no-operator yielding slightly higher loss.
This supports our choice of using SiLU at the department--expert interface, which preserves feature diversity for specialization.

\paragraph{Initialization of expert matrices.}
No.1 and No.5 in \Cref{table.ablation} compare different initialization schemes for the expert-specific matrices under the same training setup.
As mentioned in \Cref{sec:method:activation}, we replace the \textbf{\texttt{expert}} weight initialization with identity initialization (\texttt{torch.nn.init.eye\_}), which reduces the reported loss (averaged over the last 200 steps) from 2.530 to 2.504.
This result indicates that, for our factorized parameterization, initializing expert transforms with a near-identity mapping leads to a lower loss than random initialization.

\paragraph{Unsuccessful attempt of strict group top-$k$.}
We then evaluate a stricter routing variant that enforces the coverage of exactly $k_{\mathrm{dept}}$ departments per token.
Enforcing \emph{strict} department coverage ensures that each token activates exactly $k_{\mathrm{dept}}$ departments.
This avoids cases where routing concentrates on only a few departments.
However, \Cref{table.ablation} shows that this variant (No.6) yields a higher training loss than the default \emph{soft} constraint (No.1), increasing from 2.504 to 2.519.
Moreover, we observe that during training, Variant No.1 activates 3.95 distinct departments on average per token, close to the configured $k_{\mathrm{dept}}=4$, suggesting that most tokens already span multiple departments even without explicit enforcement.
This indicates that the soft group-limited top-$k$ constraint already achieves a favorable balance between efficiency and model quality, making additional hard enforcement unnecessary.

Overall, the ablation results consistently support the design choices adopted in \sys: using SiLU activation, identity initialization for expert transforms, and soft group-limited routing.
These components jointly contribute to preserving model quality while enabling the decoding-time efficiency gains analyzed in previous sections.